\documentclass[10pt,twocolumn,twoside]{IEEEtran} 

\usepackage{graphicx}
\usepackage[cmex10]{amsmath}
\usepackage{array}
\usepackage{mdwmath}
\usepackage{mdwtab}
\usepackage[font=footnotesize]{subfig}
\usepackage{fixltx2e}
\usepackage{stfloats}
\usepackage{url}

\begin{document}

\title{Path Planning Using Probability Tensor Flows}

\author{
    \IEEEauthorblockN{Francesco A. N. Palmieri \IEEEauthorrefmark{1}\IEEEauthorrefmark{2}, Krishna R. Pattipati\IEEEauthorrefmark{2}, Giovanni Fioretti\IEEEauthorrefmark{1} \\ Giovanni Di Gennaro\IEEEauthorrefmark{1}, Amedeo Buonanno}\IEEEauthorrefmark{3}\\
    \IEEEauthorblockA{\IEEEauthorrefmark{1}Dipartimento di Ingegneria, Universit\'a della Campania ``Luigi Vanvitelli'', Aversa (CE), Italy \\
    \{francesco.palmieri, giovanni.digennaro\}@unicampania.it,  \\ giovanni.fioretti@studenti.unicampania.it}, \\
    \IEEEauthorblockA{\IEEEauthorrefmark{2}Department of Electrical and Computer Engineering,
University of Connecticut, Storrs, CT, USA \\ krishna.pattipati@uconn.edu} \\
     \IEEEauthorblockA{\IEEEauthorrefmark{3}ENEA, Energy Technologies Department, Portici (NA), Italy \\ amedeo.buonanno@enea.it}
\thanks{Work partially supported by  POR CAMPANIA FESR 2014/2020, ITS for Logistics, awarded to  CNIT (Consorzio Nazionale Interuniversitario per le Telecomunicazioni). Research of Pattipati was supported in part by the U.S. Office of Naval Research and US Naval Research Laboratory under Grants \#N00014-18-1-1238, \#N00173-16-1-G905 and \#HPCM034125HQU, and by  a Space Technology Research Institutes grant (number 80NSSC19K1076) from NASA’s Space Technology Research Grants Program. } \thanks{}}

\maketitle

\begin{abstract}
Probability models have been proposed in the literature  to account for ``intelligent'' behavior in many contexts. In this paper, probability propagation is applied to model agent's motion  in potentially complex scenarios that include goals and obstacles.  The backward flow provides precious background information to the agent's  behavior, viz., inferences coming from  the future determine the agent's actions. Probability tensors are layered in time in both directions in a manner similar to convolutional neural networks. The discussion is carried out with reference to a set of simulated grids where, despite the apparent task complexity, a solution, if feasible,  is always found. The original model proposed by Attias \cite{Attias2003} has been extended to include non-absorbing obstacles, multiple goals and multiple agents. The emerging behaviors are very realistic and demonstrate great potentials of the application of this framework to real environments.   
\end{abstract}

\smallskip



\section{Introduction}

An agent's intelligent behavior is characterized by  goals and  environmental constraints in a sea of uncertainties. Probabilistic models provide a very promising  framework  for manipulating mathematically systems in which only partial knowledge is available. In a real environment,  much needs to be learned from experience starting from just a few basic structural rules.   

To be able to predict or control agents' motion in  partially structured environments, is a current challenge in a number of applications, ranging from robot planning to self-driving cars to surveillance of critical areas.  Many  methods have been proposed in the literature. See \cite{Rudenko2019} for a recent comprehensive  review. In some of our previous works, we have explored various techniques  that include  traditional Kalman filter-based approaches \cite{CosciaFusion2018} \cite{CosciaIEEE2018}, models that include social forces 
\cite{CosciaWirn2018}, polar histograms \cite{CosciaIVC2018} \cite{CosciaFusion2016} and Markov models \cite{CastaldoWirn2014}. 

In all cases, motion dynamics have to be properly combined with the environmental constraints in an attempt to capture   the agents' ``intelligent'' behavior in avoiding obstacles, in interacting with other agents and in attaining specific goals that may be dynamically changing.   

There is a growing body of literature that proposes stochastic models based on free-energy principle \cite{Buckley2017} \cite{Parr2018} \cite{Baltieri2017} and  KL-learning \cite{Zenon2018} as  general rules of intelligent behavior \cite{ParrFrontiers2018} \cite{Kaplan2018}. There are also interesting connections to causal reasoning \cite{NairSavareseetal2019}.

Since the original paper by Attias \cite{Attias2003}, there has been a growing body of literature in clarifying the connections between Markov Decision Processes  (MDPs) and probabilistic approaches  
\cite{Touissaint2006}\cite{Toussaint2009}. The objective function, typically the likelihood function, is combined with the expected reward as explained in a recent review \cite{Levine2018}.  

In this paper, we build on these findings and propose a computational framework 
in which motion and actions are modeled  jointly as probability tensors. Despite the apparent computational complexity of this characterization, motion dynamics can be handled in a manner similar to popular multi-layer neural networks  software \cite{TensorFlow2015}. Interactions among agents and obstacles in path planning are localized in space, but their consequences are treated globally in  a probabilistic model described via forward and backward messages. 

Our approach is based on tensor propagation in a  {\em Factor Graph in Reduced normal form (FGrn)} \cite{Palmieri2016}. An FGrn  is composed of the interconnection of {\em Single-Input/Single-Output (SISO)} blocks. Tensor messages are propagated bi-directionally and  combined using the sum-product algorithm \cite{Loeliger2004}. FGrn could be easily augmented to fuse heterogeneous information sources  as they become available \cite{Palmieri2016}. In this paper, we limit ourselves to discrete variables, but continuous distributions can be easily handled as in \cite{Palmieri2016}\cite{CastaldoAero2015}. Gaussian messages have been introduced in \cite{Loeliger2007a}  and used for Kalman filter tracking in \cite{Palmieri2016}\cite{CastaldoAero2015}.   More details about factor graphs  can be found in the seminal papers  \cite{Loeliger2004} and \cite{Loeliger2007a}. Further developments on the factor graphs in the reduced normal form are in \cite{Palmieri2016}. 

In this paper, we explore an application in which states are defined on a 2D  finite grid with the additional dimension utilized for actions. This extension allows the distributions on the map to account  for intended motion directions towards one or more goals. Interaction with obstacles are included in the model and the probability flow travels in time in both directions. Forward and backward probability diffusion resembles convolutional operations in multi-layer  neural networks. 

In this work, the presence of obstacles is modeled as censored/renormalized stochastic conditional distributions,  that keep the agent away from obstructions. Therefore, obstacles are not seen as absorbing states \cite{Attias2003}, but rather as random reflectors that redistribute the probability flow on the map.
We will see how the backward  flow plays a crucial role in the probabilistic model.  Information ``coming from  the future''  may be seen as inverse dynamic modeling, or like a probability field diffused in the environment, or like social interactions. This is a mathematical translation of the fact that intelligent agents generally base their decisions on the impact of projected actions into the future.   

   Time also plays a crucial role as generally the agents would like to reach their goals in the shortest time possible. We show how minimum time determination is straightforward using the backward flow and how this technique will always be able to find the best feasible paths: the map can be an arbitrarily very complicated maze of regions. 
 
 In this paper, we first extend the basic one-agent/one-goal model to include multiple goals: the probabilistic framework allows the inclusion of a distributed  set of targets, by the introduction of spread-out  probability values  at the end of the chain in the backward flow.  Furthermore, we extend the model to include multiple agents: each agent is driven by a different probability flow and interacts with the other flows by seeing the other agents as moving obstacles, or as targets.  The probability distributions  are dynamically updated  and the actions for each agent  are computed accordingly. We are not aware of similar extensions for such potentially complex scenarios  in the literature. 
 
In Section \ref{sec:bayes}, we review the basic Bayesian model and its application to path planning.  In Section \ref{sec:trans}, we present the state transition model for obstacle avoidance and in Section \ref{sec:control}, we discuss the action-control mechanism, that can go from a simple diffusion to the maximum reward policy. The discussion is carried out with reference to various finite grids. Computational complexity and memory requirements are  included in Section \ref{sec:complexity} and  relations to dynamic programming are provided in Section \ref{sec:mdp}. Extensions to multiple goals are in Section \ref{sec:multiple} and the application to more complex multiple-agent tasks is in Section \ref{sec:many}. Section \ref{sec:conc} discusses conclusions and future developments.

\section{The  Bayesian Model}
\label{sec:bayes}

\begin{figure}[ht]
\centering
\includegraphics[width=8cm]{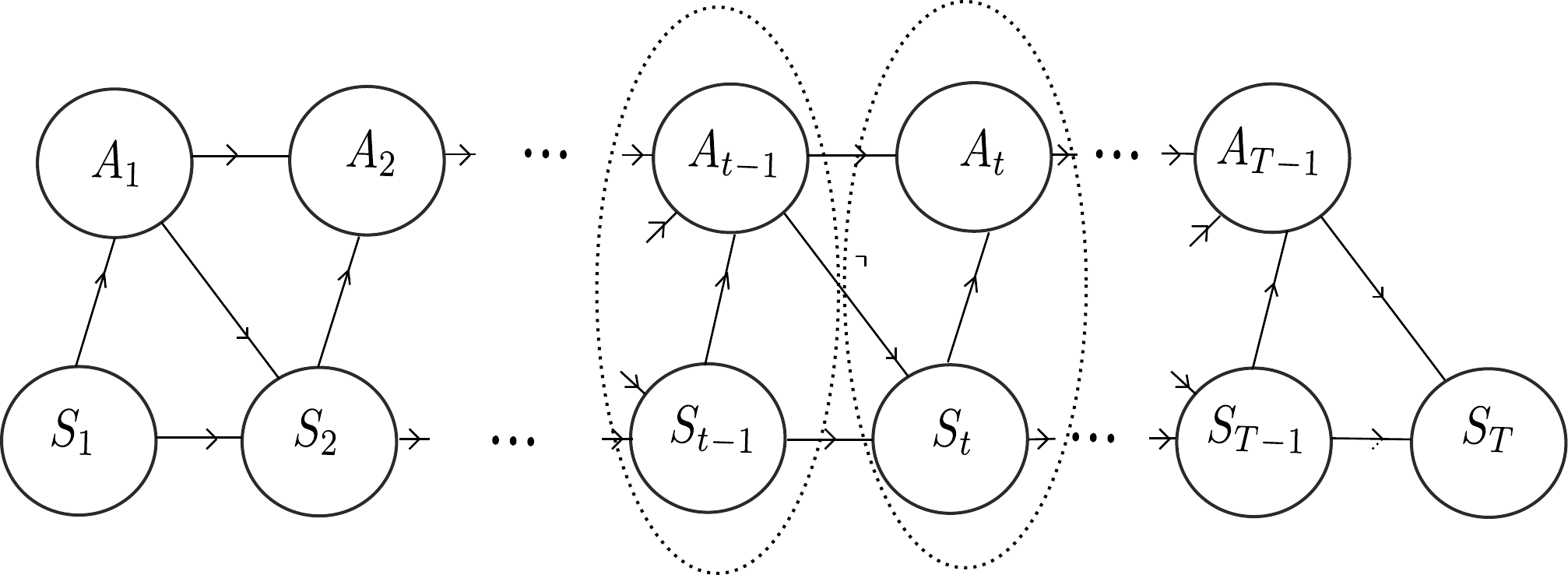} 
\caption{State-Action Markov Model as a Bayesian graph}
\label{fig:MM}
\end{figure}

Our model for a single agent is based on a sequence of states $S_t$, $t=1:T$, that  are assumed to be fully observable and  to belong to a finite space ${\cal S}=\{s^1,...,s^{n_S}\}$ and  a sequence of ``actions'' $A_t$, $t=1:T-1$, that also belong to a finite space ${\cal A}=\{ a^1,...,a^{n_A}\}$. In a general Markov  model, both state and actions at time $t$ are linked to both state and action at time $t-1$. 
The stochastic model is fully characterized by the pmfs (probability mass functions)\footnote{We use upper case letters for random variables and lower case letters for their values.}
\begin{equation}
\left\{ \begin{array}{l} 
\pi_{S_1A_1}(s_1,a_1); \\
p_{S_tA_t|S_{t-1}A_{t-1}}(s_t,a_t|s_{t-1},a_{t-1}), t=2:T-1;\\
p_{S_T|S_{T-1}A_{T-1}}(s_T|s_{T-1},a_{T-1}).
\end{array} \right.
\label{eq:general}
\end{equation}
 To be more specific about mutual connections, let us rewrite the general pmf as 
\begin{equation}
\begin{array}{l}
p_{S_tA_t|S_{t-1}A_{t-1}}(s_t,a_t|s_{t-1},a_{t-1}) \\
={  p_{S_tA_tS_{t-1}A_{t-1}}(s_t,a_t,s_{t-1},a_{t-1})  \over p_{S_{t-1}A_{t-1}}(s_{t-1},a_{t-1}) } \\
={  p_{A_t|S_tS_{t-1}A_{t-1}}(a_t|s_t,s_{t-1},a_{t-1})  p_{S_tS_{t-1}A_{t-1}}(s_t,s_{t-1},a_{t-1}) \over p_{S_{t-1}A_{t-1}}(s_{t-1},a_{t-1}) } \\
=  p_{A_t|S_tS_{t-1}A_{t-1}}(a_t|s_t,s_{t-1},a_{t-1})  \\
\hspace{1 in} \cdot p_{S_t|S_{t-1}A_{t-1}}(s_t|s_{t-1},a_{t-1}). 
\end{array}
\end{equation}
Now, if we drop the conditional dependence of $A_t$ on $S_{t-1}$, we have the factorized model 
\begin{equation}
\begin{array}{l}
p_{S_tA_t|S_{t-1}A_{t-1}}(s_t,a_t|s_{t-1},a_{t-1}) \\
~~~~~=p_{A_t|S_tA_{t-1}}(a_t|s_t,a_{t-1}) p_{S_t|S_{t-1}A_{t-1}}(s_t|s_{t-1},a_{t-1}),
\end{array}
\end{equation}
depicted in Figure \ref{fig:MM}. 

Note that the factor $p_{S_t|S_{t-1}A_{t-1}}(s_t|s_{t-1}a_{t-1})$ represents the physical system (plant), i.e., how the state changes according to the previous state and action.  The factor $p_{A_t|S_tA_{t-1}}(a_t|s_t,a_{t-1})$ is  the rule (controller), in general stochastic, with which new actions are generated as a consequence of the current state and the previous action. This is a standard stochastic model used in MDPs (Markov Decision Processes) \cite{Bertsekas2019}\cite{Puterman2005}, 
where $p_{A_t|S_tA_{t-1}}$, $i=1,...,T-1$, is also usually expressed as a {\em policy} {\large \boldmath$\pi$}. Action policy will be introduced in a following discussion also because in our formulation,  $(A_t,S_t)$ will be treated together by the control action.  

\begin{figure*}[ht]
\centering
\includegraphics[width=17cm]{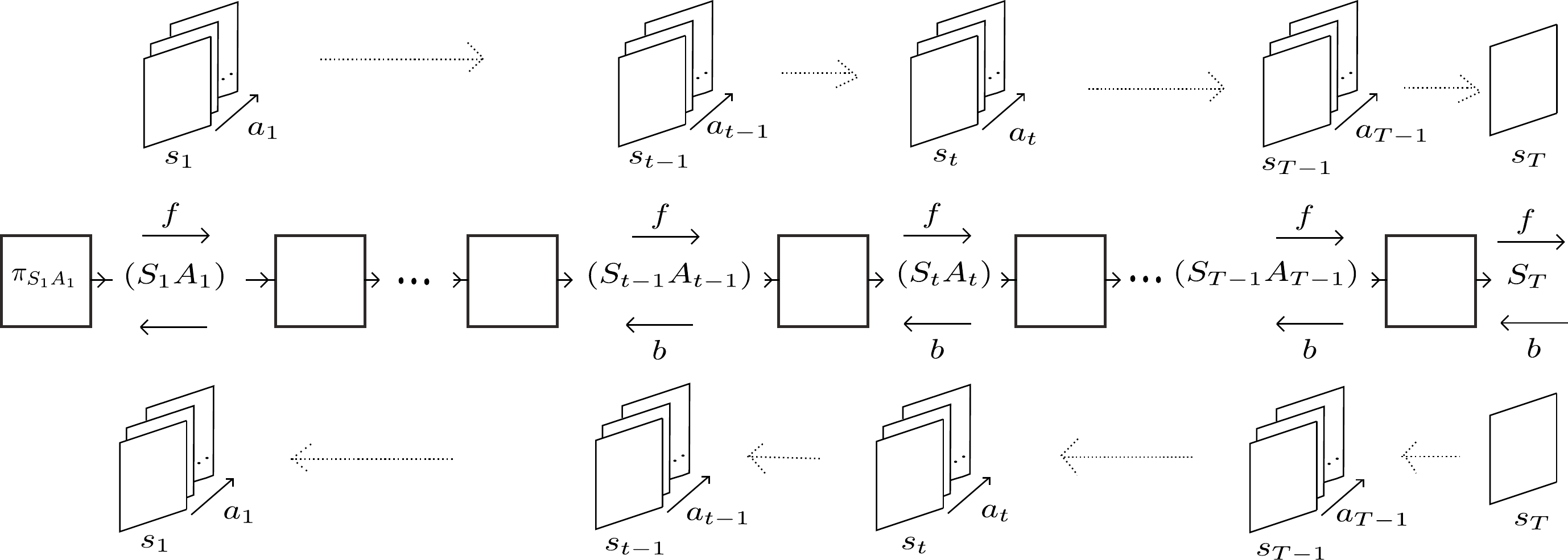} 
\caption{State-Action Markov Model as a Factor Graph with forward (top) and backward (bottom) tensor flows.}
\label{fig:FGrn}
\end{figure*}

Recognizing the cliques in Figure \ref{fig:MM}, we can reduce the system to the Markov chain depicted in the middle of Figure \ref{fig:FGrn} as a Factor Graph. Belief propagation can be performed on the graph with states and actions at each time $t$ considered jointly. Therefore, all forward ($f$) and backward ($b$) messages  are defined on the $n_S \times n_A  $ dimensional space ${\cal X}\times {\cal A}$. 

Using the shortened notation, forward and backward  messages are composed using the sum rule \cite{Loeliger2004}\cite{Palmieri2016}
\begin{equation}
\begin{array}{l}
 f(s_1,a_1)=\pi(s_1,a_1); \\
f(s_t,a_t)= \\ 
\hspace{0.1 in} \sum_{a_{t-1}}  p(a_t|s_t,a_{t-1}) \sum_{s_{t-1}}  p(s_t|s_{t-1}a_{t-1})
  f(s_{t-1},a_{t-1})  \\ \hspace{1.5 in} t=2:T-1 ; \\
   f(s_T)= \sum_{a_{T-1}} \sum_{s_{T-1}}  p(s_T|s_{T-1},a_{T-1})  f(s_{T-1},a_{T-1}); \\
    b(s_{T-1},a_{T-1}) \propto \sum_{s_T} p(s_T|s_{T-1} a_{T-1} )b(s_T); \\
b(s_{t-1},a_{t-1}) \propto   \\ 
 \hspace{0.1 in}\sum_{a_t}  p(a_t|s_t,a_{t-1}) \sum_{s_t}  p(s_t|s_{t-1}a_{t-1})
 b(s_t,a_t) \\ \hspace{1.5 in} t=T-1:2  
\end{array}
\label{eq:prop}
\end{equation}
Posterior distributions, are obtained with the product rule \cite{Loeliger2004}\cite{Palmieri2016}
\begin{equation}
\overline{p}(s_t,a_t) \propto  b(s_t,a_t) f(s_t,a_t).
\end{equation}
Note that the forward messages, are  normalized distributions, if  $\pi(s_1,a_1)$ is normalized, while the backward messages and  the posteriors are only proportional to their respective  pmfs. In belief propagation, messages and posteriors can  be kept unnormalized even if it is preferable to normalize them for numerical stability \cite{Loeliger2004}\cite{Palmieri2016}. The reader not too familiar with probability propagation should be aware that these rules are  rigorous translations of Bayes' theorem and marginalization. 

Our discussion will be carried out with reference to a 2D scenario in which the states are defined on  
an $N \times M$ finite grid
\begin{equation} 
S_t=(X_t,Y_t), ~~~~~X_t\in \{1,...,M\};~~Y_t\in  \{ 1,...,N\}, 
\end{equation}
$n_S=N\cdot M$. In our examples, there are nine possible actions:  
\begin{equation}
\begin{array}{l}
{\cal A}=\{ a^1,...,a^9\}=\{{\rm still, up,  up-right, right,  } \\
 \hspace{0.4 in}{\rm down-right, down, down-left,left,up-left   }\}
 \end{array}
\end{equation}
Hence, all the distributions  are tensors in which the first two dimensions are the coordinates on the grid and the third dimension is the action (Figure \ref{fig:FGrn}, top and bottom).   

\section{The State-transition Model}
\label{sec:trans}

\begin{figure}[ht]
\centering
\includegraphics[width=6cm]{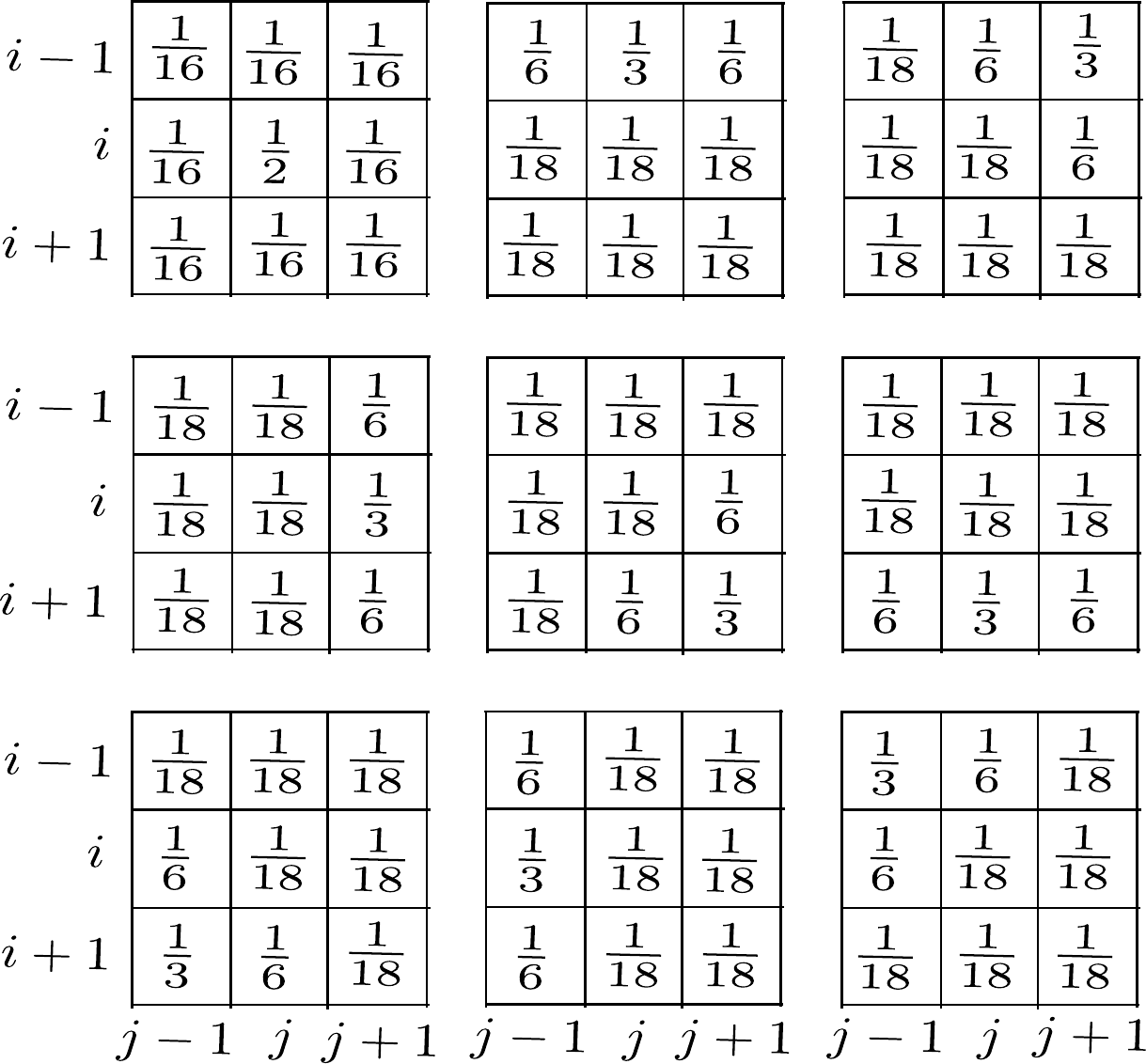} 
\caption{State transition distributions for $(X_t,Y_t)$ conditioned on the previous state being in  
$(x_{t-1},y_{t-1})=(i,j)$  and $a_{t-1}=$still, up, up-right, right, down-right, down, down-left, left, up-left (in lexicographic order).}
\label{fig:masks}
\end{figure}

The probability function  $p_{S_t|S_{t-1}A_{t-1}}(s_t|s_{t-1}a_{t-1})$ describes the system state transition at time $t$  as a consequence of the previous state and action. In our 2D implementation, in the absence of obstacles, the stochastic motion is  described in Figure \ref{fig:masks}, where  the distribution is limited to one step away from the previous position and it is mostly concentrated in the direction of the intended motion. The probabilities shown in the figure are only a possible choice. The distribution could be sharper to reflect more determined action in each direction, or be smoother for an agent that moves more randomly. Furthermore, larger masks could be easily defined to model agents that may move more than one step at a time. 

The inclusion of obstacles and boundaries (obstructions) in the definition of  $p_{S_t|S_{t-1}A_{t-1}}(s_t|s_{t-1}a_{t-1})$ is crucial to obstacle avoidance and best path selection strategies. 
Boundaries and obstacles are defined on the state grid with an $N \cdot M$ binary mask $W=[w_{ij}]$, with $w_{ij}=1$, if there is an obstacle in $(i,j)$ and $w_{ij}=0$ for free locations.

\begin{figure}[ht]
\centering
\includegraphics[width=4cm]{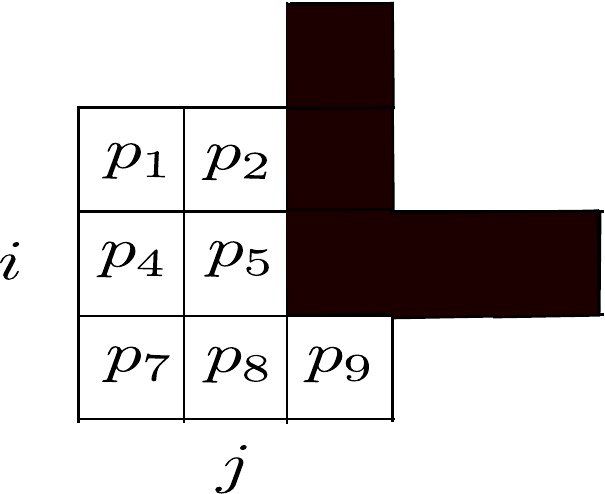} 
\caption{Example of state transition renormalization in the presence of an obstruction. We are at time $(t-1)$ and  $p_1,...,p_9$ would be the probabilities  $p((x_t,y_t)| (x_{t-1},y_{t-1})=(i,j), a_{t-1}=\bar a)$ for a specific action $\bar a$ (one of the masks in Figure \ref{fig:masks}) if we had no obstruction. Since pixels $(i-1,j+1)$ and $(i,j+1)$ are obstructed, $p((x_t,y_t)| (x_{t-1},y_{t-1})=(i,j), a_{t-1}=\bar a)$ is redefined with $p_3,p_6 \leftarrow 0$ and $p_i \leftarrow {p_i \over 1- p_3-p_6}$, $i=1,2,4,5,7,8,9$.}
\label{fig:obst}
\end{figure}

In our implementation, we have assumed that the obstacles  are avoided, with  a mechanism that censors and renormalizes the transition distribution as shown in the example  of Figure \ref{fig:obst}: when the transition distribution overlaps an obstacle, the overlapping probabilities are set to zero and the others are renormalized.  This means that the agent is reflected (discouraged) by the obstacles and is projected onto the non-overlapping pixels with the same  probabilities, but renormalized. 

\begin{figure}[ht]
\centering
\includegraphics[width=8cm]{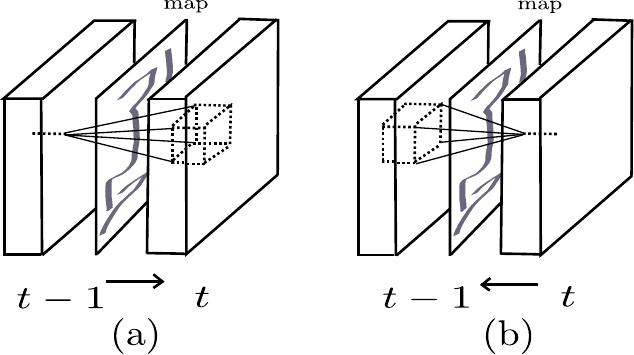} 
\caption{Diffusion in the forward (a) and in the backward flow (b).}
\label{fig:conv}
\end{figure}

Figure \ref{fig:conv} shows the localized operations in the forward and in the backward flow. They  resemble a layer in a convolutional multi-layer network. The difference here is that the convolutional filters are not space-invariant as they depend on the map. Furthermore, the coefficients are probabilities and there are possible normalizations. A direct correspondence to a multi-layer neural network is an intriguing problem and it is currently under investigation.  

\begin{figure}[ht]
\centering
\includegraphics[width=4cm]{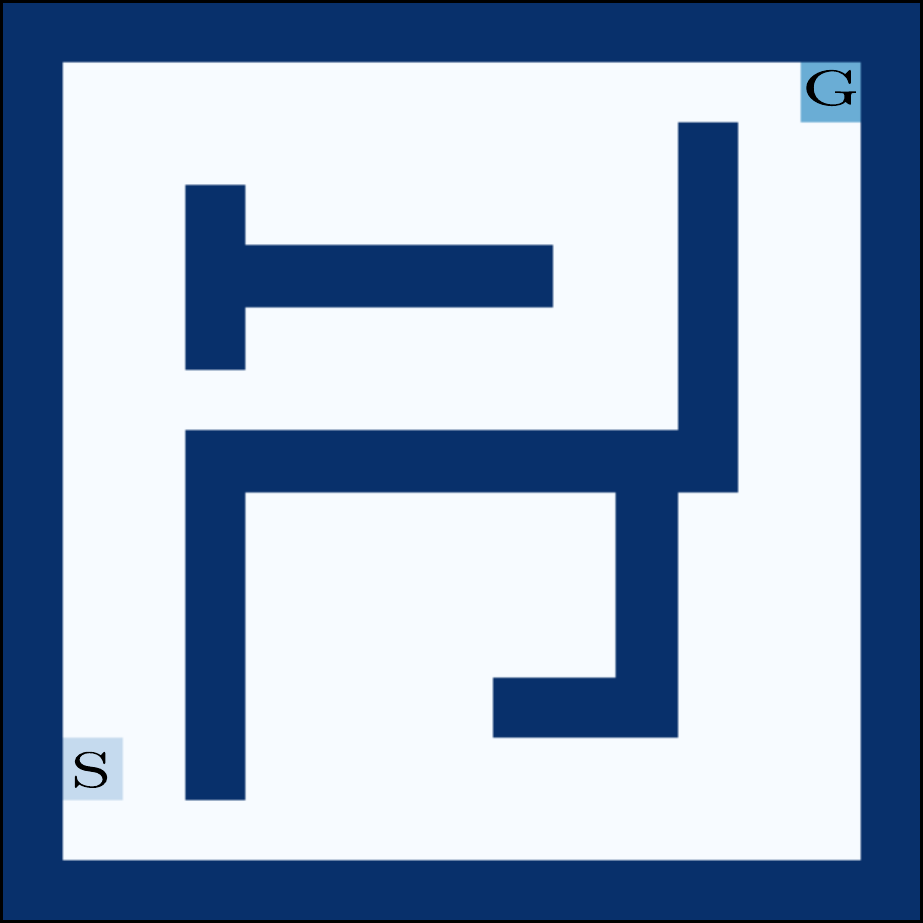} 
\caption{A $15 \times 15$ grid with Start (S), Goal (G), obstacles and boundaries.}
\label{fig:scene}
\end{figure}

\vspace{-0.8cm}
\section{The Control Module}
\label{sec:control}

Action generation at time $t$  is described by the pmf  $p(a_t|s_t,a_{t-1})$ that models the new action as a consequence of previous action $a_{t-1}$ and current state $s_t$. This is the crucial control part of the agent's behavior that, in a complete scenario, should depend also on projected inference into the future and the goal via a certain policy {\large \boldmath$\pi$} as in MDP \cite{Bertsekas2019}\cite{Puterman2005}\cite{Touissaint2006}. 

\subsection{Pure diffusion}

We look first at a simplified scenario in which the agent moves in a pure diffusion mode driven by  the pmf $p(a_t|a_{t-1})$, independently from the state and keeping only possibly a memory on the previous action $a_{t-1}$. This action memory allows the modeling of  motion stiffness, i.e. a tendency of the agent to maintain its course,  or in general to maneuver according to a Markov process (with no relation to the state).  Since we have $n_A$ possible actions (in our example $n_A=9$), $p(a_t|a_{t-1})$ is described by an $n_A \times n_A$  row-stochastic matrix  $P_A$ .  
 
 In this first analysis of our model, we assume that initial (start) and the final (goal) state are fixed (Attias' model \cite{Attias2003}) (in Bayesian networks' terms we say that the states $S_1=\overline s$ and $S_T= \overline g$ have been {\em instantiated}). This corresponds to messages at the beginning and at the end of the chain 
 \begin{equation}
 f_{S_1A_1}(s_1, a_1)= \delta(s-{\overline s}) \pi_{A_1}(a_1); \hspace{0.1 in} b_{S_T}(s_t)=\delta(s-{\overline g}),
 \end{equation}    
where $\pi_{A_1}(a_1)$ is the initial action distribution. If we set a deterministic initial action $a_1=\bar a$, we  have $\pi_{A_1}(a_1)=\delta(a_1-\bar a)$, where $\delta(x)=1$, for $x=0$ and zero else. 

\begin{figure}[ht]
\centering
\includegraphics[width=8cm]{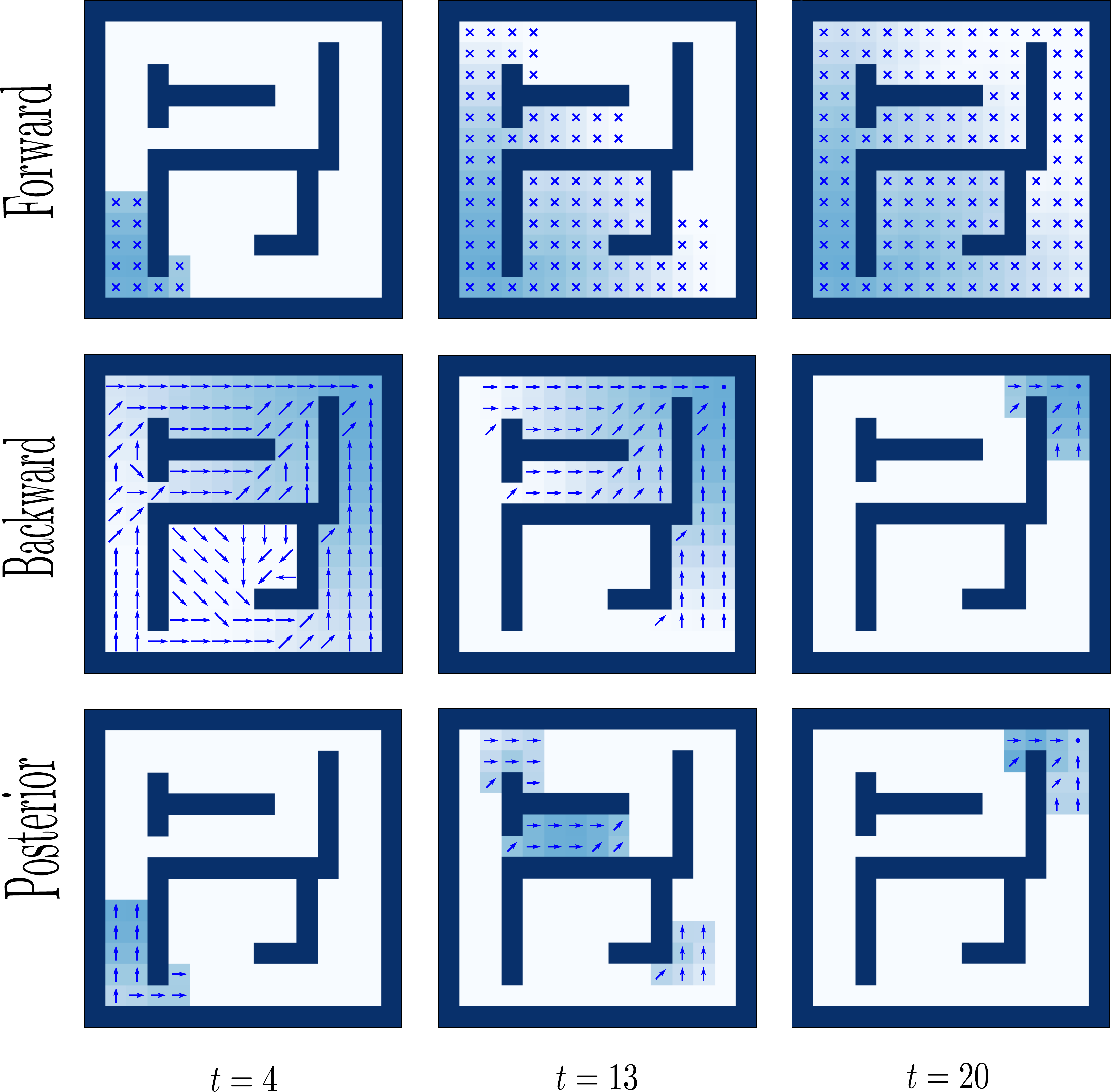} 
\caption{Forward, backward and posterior distributions at time steps $t=4,13,20$, in a pure diffusion mode. State probabilities, marginalized over actions,  are shown at different locations in shades of blue. Action probabilities are shown on each location as arrows pointing in the direction of maximum probability. A dot denotes that the maximum probability is for ``still''.  Crosses denote uniform action distributions. The time horizon is $T=23$ (not minimum).}
\label{fig:diff}
\end{figure}

Figure \ref{fig:scene} shows a $15 \times 15$ grid, in which an agent from a starting point S, has to find his way through a set of obstacles to reach the goal G. Boundaries are considered obstacles and handled with the reflection mechanism explained in Figure \ref{fig:obst}.  The time horizon is $T=23$. Initial action distribution is set to uniform and the action transition matrix $P_A$ is uniform ($1/9$ in all entries).  

Figure \ref{fig:diff}  shows the results of a pure diffusion on the  grid of Figure \ref{fig:scene}.
The three rows show forward, backward and posterior distributions at time steps $t=4,13,20$. Recall that each distribution is three-dimensional and action is in the third dimension. State probabilities (marginalized over actions), are shown  proportional to the pixel intensity and the action probabilities are shown as arrows in each pixel: each arrow points in the direction of maximum action probability. Furthermore  pixels with a dot correspond to maximum action probability on ``$still$''.  Crosses denote uniform action distributions.

Note how the forward distribution is a pure diffusion with no preferred direction. Backward distributions are instead  ``inverse diffusion'' starting from the goal at time $T$ and working their way back into the past. Posterior distributions, that are the normalized product of forward and backward messages, show the most likely  regions and directions in time  compatible with the agent reaching the goal at time $T$.  The crucial aspect of this analysis is the presence of the action dimension that gives direction to the motion.  The distribution on  position only on the grid, would not give us complete information on how the probability flow progresses.  

Note how the backward flow resembles a vector field suggesting intriguing relations between this stochastic model and traditional field theory. It is as if the backward flow was leaving tracks on the grid (``a yellow brick road'')  that must be followed to reach the goal in the set time.  We have verified that no matter how complicated the maze is, if a feasible path exists, the combination of forward and backward diffusion always concentrates the probability mass in the appropriate regions of the space, where the agent should be located in time to reach the goal.

The main limitation of the pure diffusion model, is that the time it takes to reach the goal has to be pre-determined \cite{Attias2003}\cite{Touissaint2006}. If $T$ is too small, there exist no regions where  the agent can be found that are compatible with the agent reaching the goal: the posterior probability is zero because forward and backward flows are always such that one of them is zero. If $T$ is too large, the agent can span the space with more freedom and there are many (sub-optimal) ways of reaching the goal. 

\begin{figure}[h]
\centering
\includegraphics[width=8cm]{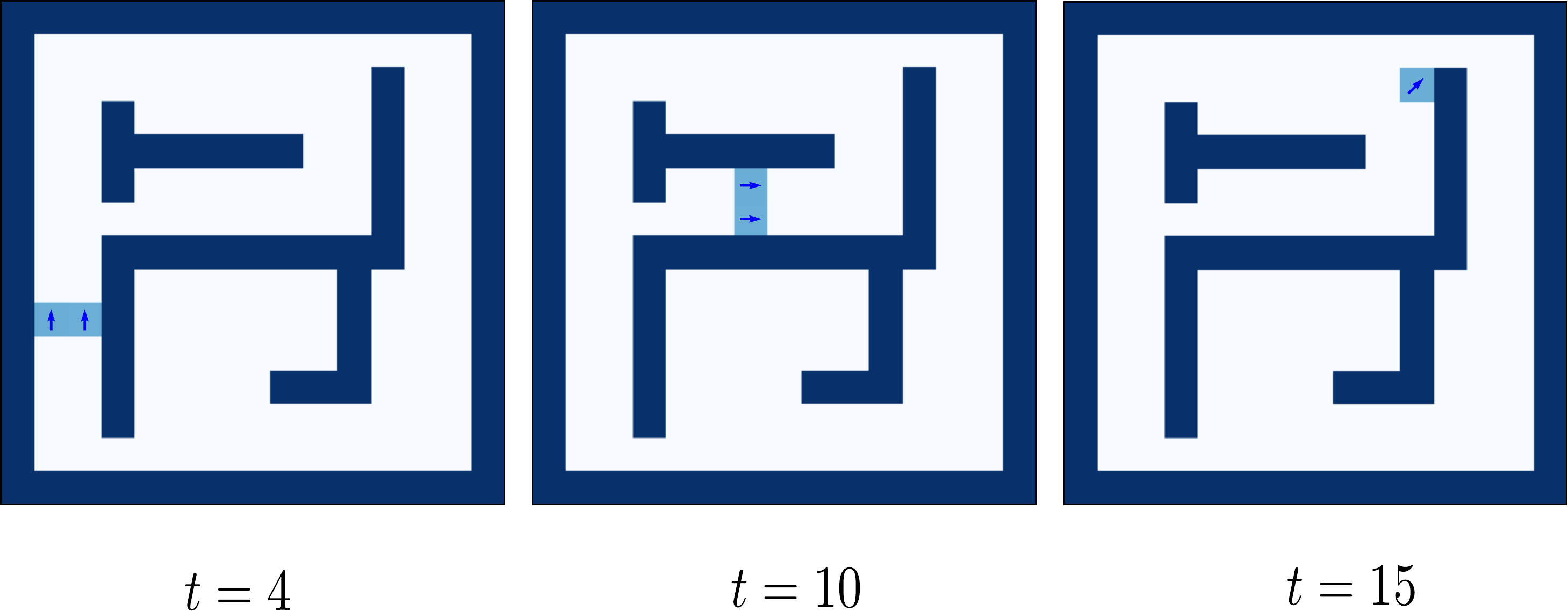} 
\caption{Posterior distribution for  $T=T_{min}=18$, $t=4,10,15$}
\label{fig:post}
\end{figure}

Figure \ref{fig:post} shows the posterior distribution at three time steps, when the time horizon has been set to exactly the minimum $T=T_{min}=18$.  There are only small regions where the agent may be found at any given time. Only those locations are compatible for the agent to reach the target with a minimum length path.  This is to be compared with the posterior in Figure \ref{fig:diff} where the time horizon has been set to a non-minimum value ($T=23$), for which there are wide regions where the agent may be located at any given time. 

\begin{figure}[h]
\centering
\includegraphics[width=8cm]{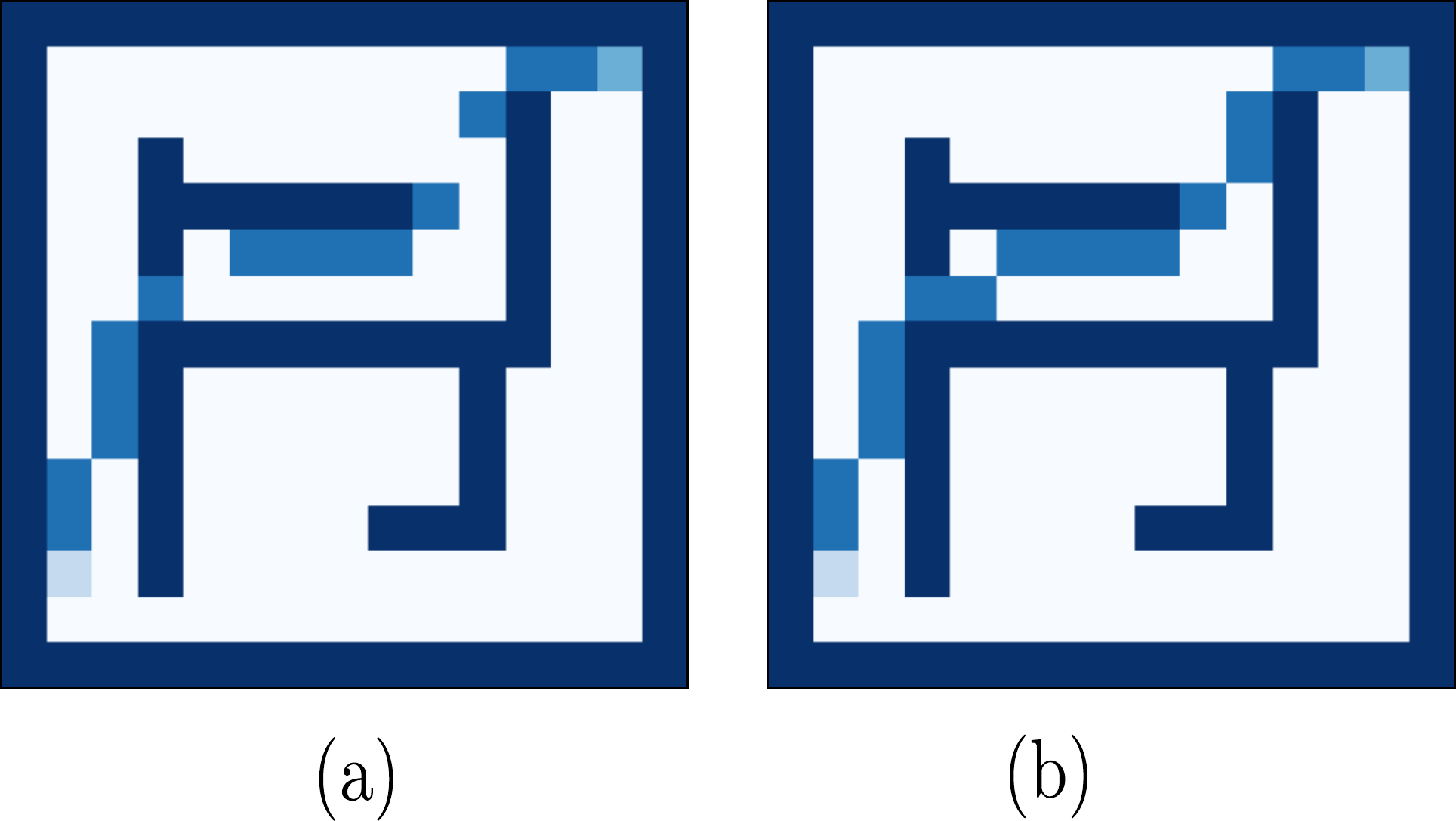} 
\caption{(a) The set of maximum posterior points for $T =23> T_{min}$ (note the disconnected path); (b) The set of maximum posterior points for $T=T_{min}=18$.}
\label{fig:maxp}
\end{figure}

\begin{figure}[h]
\centering
\includegraphics[width=8cm]{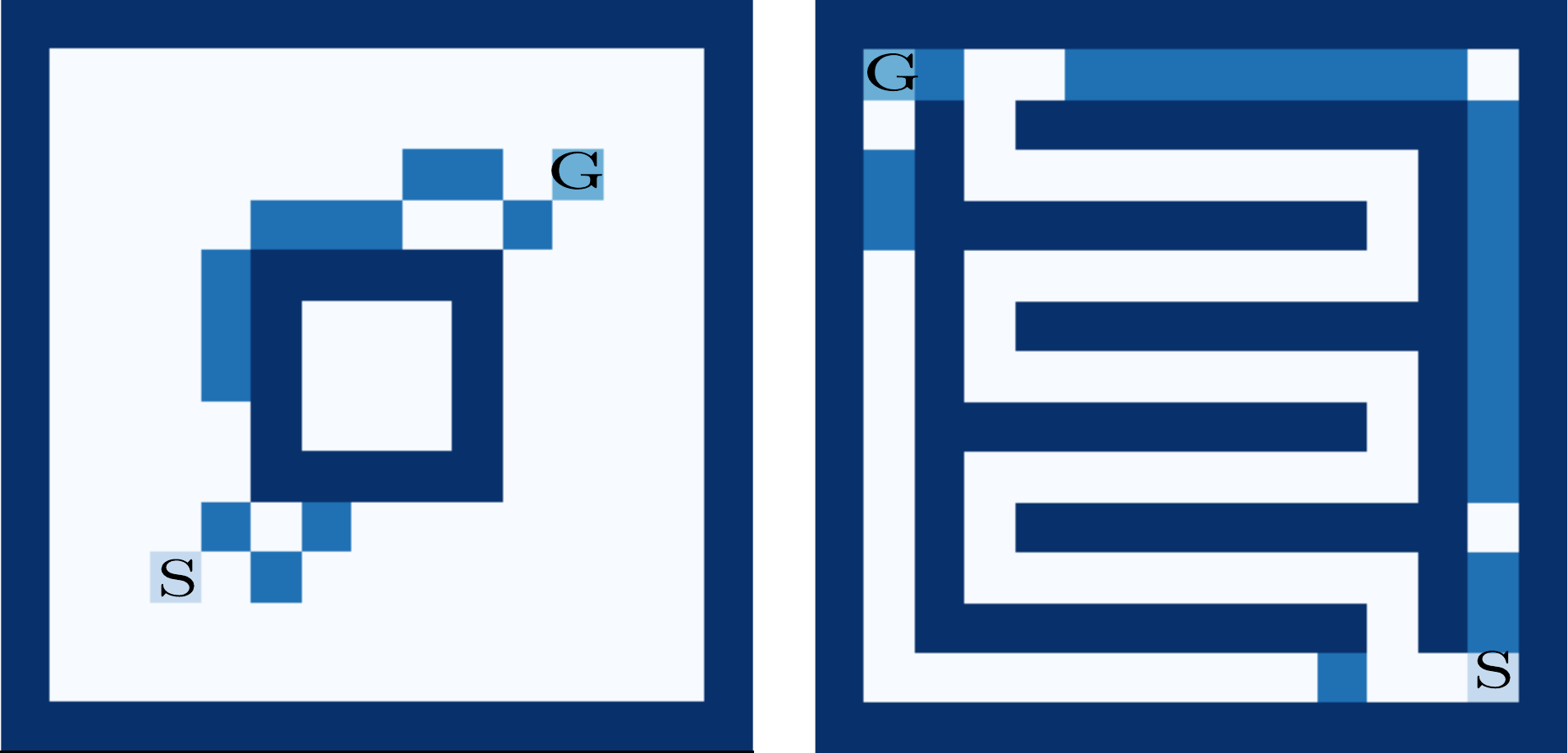} 
\caption{(a) The set of maximum posterior points for $T=T_{min}$ on two grids where the path results disconnected. }
\label{fig:maxp12}
\end{figure}
\subsubsection{Path determination}

Maximization of the total likelihood would suggest that, to determine the path to destination, we just have to compute the points of maxima in the posterior distribution. Unfortunately, this unconstrained solution may provides disconnected paths as shown in  Figure \ref{fig:maxp}(a).  When a non-minimum time horizon is chosen, the agent has more time to reach the goal: it can be  more freely located in larger areas where the posterior distribution may take various configurations. We have also verified that using the means, rather than the maxima, would not change much. Figure \ref{fig:maxp}(b) shows the results of choosing minimum time. In this case, the more constrained posterior support  produces a connected path. Unfortunately, even with minimum time, we are not guaranteed that the path with the maxima points is connected. 
Figure \ref{fig:maxp12} shows two more examples where, even if the time horizon is set to the minimum value, we obtain disconnected paths. This effect seems to be more pronounced  when we have multiple minima.

\subsubsection{Minimum time determination}

Even if a pure diffusion model (no action control) cannot guarantee connected paths, it may be useful to compute minimum time. This can save time in the computations and gives us a frame of reference for the scenario with control included that follows. The minimum time can be computed through probability diffusion in a straightforward way using the backward flow:  {\em run the backward flow until at the start position there is a non-zero probability}. This is another nice feature of using probability propagation. 
    
\subsection{Introducing state-dependent actions}

Path determination in general should not depend on the minimum time determination. More specifically, it should be determined using an algorithm that fixes the state and the action at every time step and then takes an informed decision to evolve to the next step. This is what the state-dependent action discussed in this section does.  
    
The probabilistic model shown above demonstrates that the interaction between the forward and the backward information flow can provide a framework for solving an apparently very complicated problem. The great feature of this model, is that the backward flow, that corresponds to running the agents' dynamics backward in time, can provide a nice track for guided behaviour. 
\begin{figure*}[ht]
\centering
\includegraphics[width=17cm]{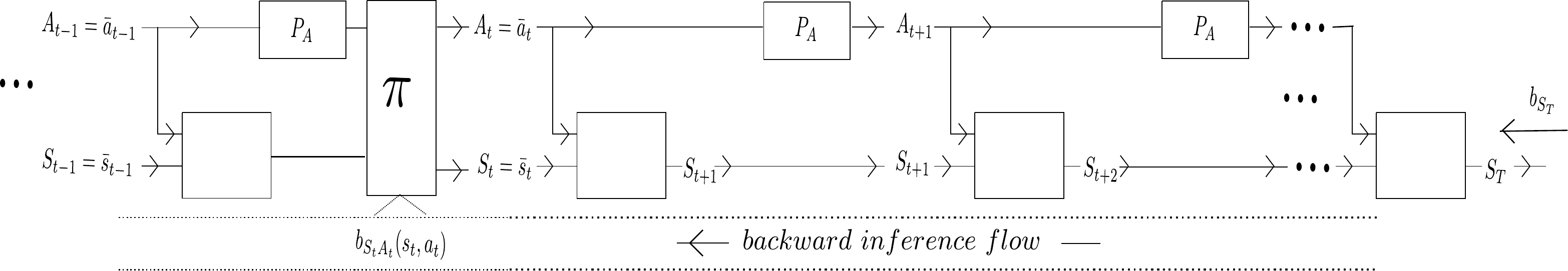} 
\caption{Block diagram of the plant-controller system at time $t$. The backward message at $T$ is $b_{S_T}(s_T)= \delta(s_T- {\bar s}_T)$.}
\label{fig:plant}
\end{figure*}
Figure \ref{fig:plant} is a block diagram of the system at time $t$,  that it is now a hybrid architecture that mixes probabilistic inferences with decisions. The system has to generate a single path with a sequence of specific instances $\{ {\bar s}_t,{\bar a}_t\}_{t=1}^T$ with actions and states  now generated using  also  backward information coming  from the future.  The agent is driven by an algorithm that uses goal diffusion ``rolled''  back from the future  in a way  similar to the case of pure diffusion. 

Figure \ref{fig:plant} shows the situation at a generic time $t$, where at previous time steps decisions on states and actions have been taken.  At time $t-1$, we have an instance  $(S_{t-1},A_{t-1})=({\bar s}_{t-1},{\bar a}_{t-1})$ and the controller follows a policy {\large \boldmath$\pi$} to set $(S_t,A_t)=({\bar s}_t,{\bar a}_t)$. Note that backward information from the future is based on pure diffusion: no specific control action is projected into the future. Nevertheless the backward distributions provide guidance to choose appropriate state and actions  towards  regions of larger  probabilities. Note also that {\em the action policy  treats $(A_t,S_t)$ jointly}.

The following algorithm follows a greedy strategy and finds a connected path. 

\smallskip
\noindent
{\em G-algorithm Outline:} 

\noindent
(a) Initialize state  (goal) at time $t=T$ with  $b_{S_T}(s_t)=\delta(s-{\overline g}) $; 

\noindent
(b) Compute the complete backward flow using the last two equations in (\ref{eq:prop}) from $t=T$ to $t=2$;

\noindent
(c) Initialize the state and action (start) at time $t=1$ with 
$f_{S_1A_1}(s_1, a_1)= \delta(s_1-{\overline s},a_1-\bar a)$; 

\noindent
(d) Set $ t \leftarrow t+1$ and compute the forward distribution $f_{S_tA_t}(s_t,a_t)$ using the second equation in (\ref{eq:prop}); 

\noindent
(e) Compute the posterior distribution:

$p_{S_tA_t}(s_t,a_t) \propto f_{S_tA_t}(s_t,a_t) b_{S_tA_t}(s_t,a_t)$; 

\noindent
(f) Set $(S_t,A_t)=({\bar s}_t,{\bar a}_t )= {\rm argmax} ~ p_{S_tA_t}(s_t,a_t)$

\noindent
(g) Replace forward distribution as:
 
$f_{S_{t}A_{t}}(s_{t},a_{t})=\delta(s_{t}={\bar s}_{t},a_{t}={\bar a}_{t})$

\noindent
(h) Back to (d) for the next time step. 

\smallskip
The algorithm essentially maximizes the likelihood in a greedy fashion. The solution is reached much like in the Viterbi algorithm (poor man's Viterbi) in a large-dimensional space. Therefore, if the time horizon is set to minimum time, the solution is optimal. 

\begin{figure}[h]
\centering
\includegraphics[width=8cm]{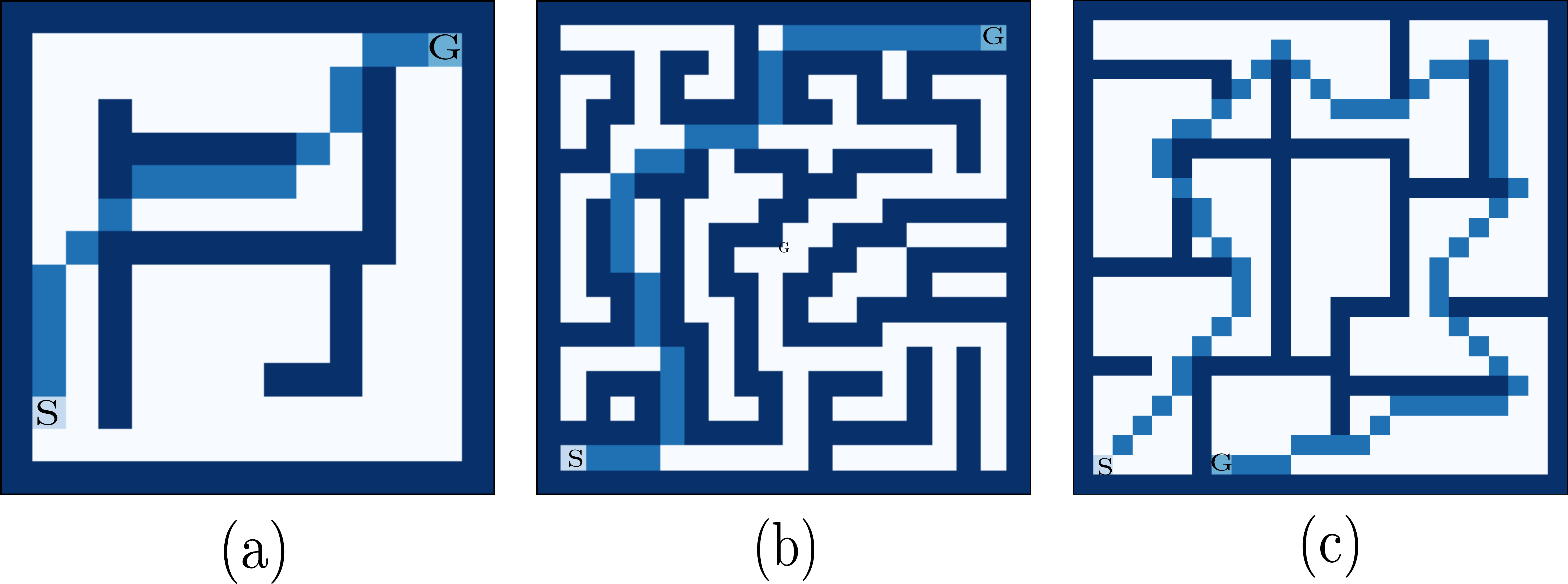} 
\caption{Path found by the G-algorithm for: (a) the grid of Figure \ref{fig:scene}; (b) (c) two more  complex grids.}
\label{fig:3paths}
\end{figure}

Figure \ref{fig:3paths}(a) shows the path resulting from the above G-algorithm for the grid of Figure \ref{fig:scene}. We have verified that, no matter how complicated the maze is, the procedure always finds a connected path to the destination if a feasible time $T$ has been set. Figure \ref{fig:3paths}(b) and (c)  show the paths found for two more grids $20 \times 20$ and $25 \times25$, respectively. 
For the time horizon, we have used the minimum time for all ($T_{min}=18, 32, 64$ respectively). 
When the time is not minimal, the path may oscillate around its final destination, but there is always a non-null probability on the goal.

\subsubsection{Algorithm variations}

In  the G-algorithm, a number of small variations  may be adopted for  more robust implementations, or for different applications: 

1. When the algorithm runs on non-minimum time, a control to stop the evolution can be added when the agent steps on the goal. This would avoid that the agent keeps wandering around and that  the path increases its  length.   

2. During evolution, at specific time $t$ and at  locations around the current agent's position,  it may happen that the posterior is null. This may be caused by an insufficient time horizon $T$, or because there is no feasible solution (forward and backward flows do not meet).  In such cases, for a single agent, the algorithm will give no answer. As we will see in the following, when there are multiple agents, i.e. the map is dynamically changing, we may adopt different strategies. For example, a random action could be taken according to the forward distribution, or simply wait for a solution to become feasible.   

\begin{figure}[h]
\centering
\includegraphics[width=8cm]{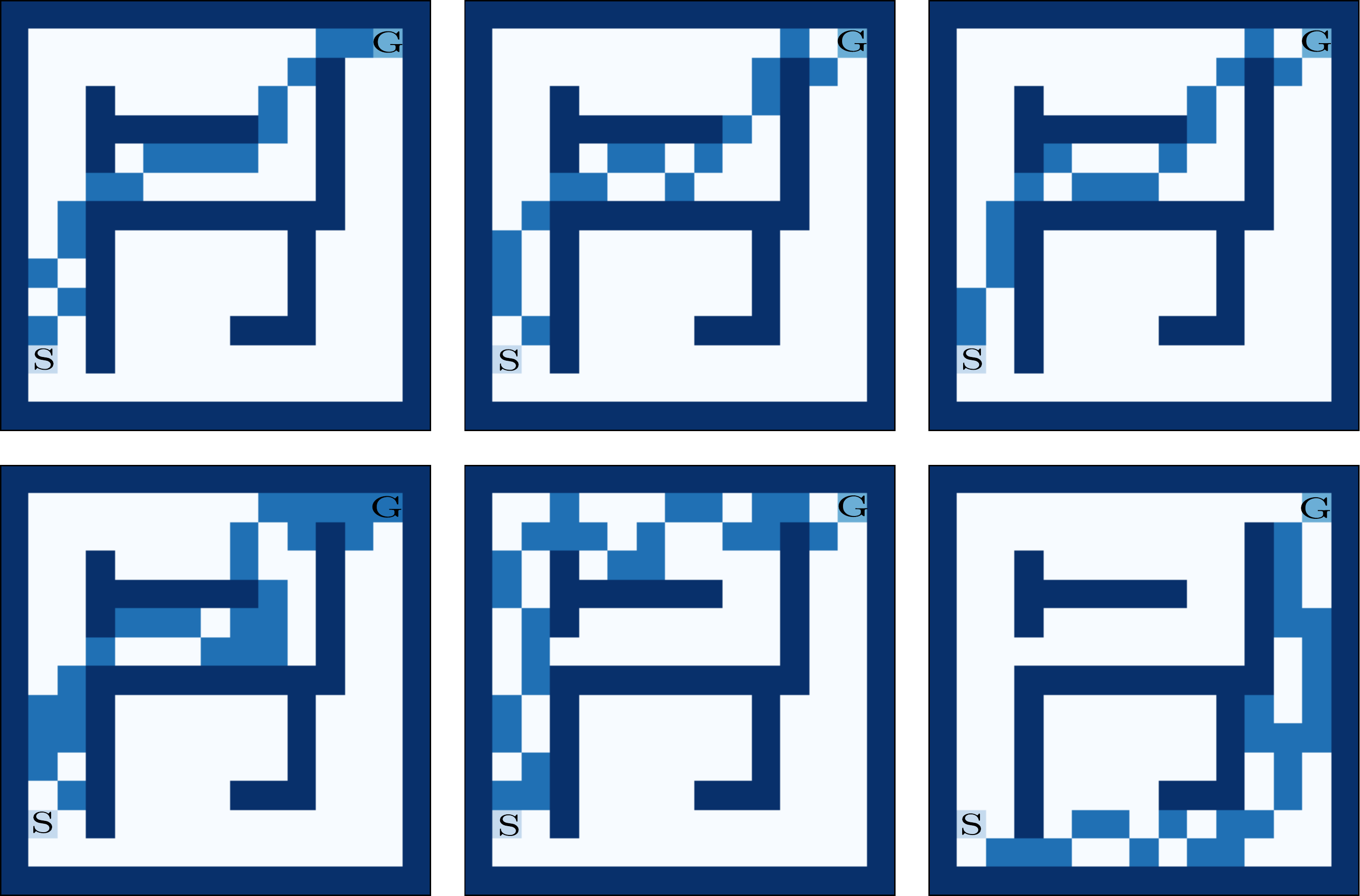} 
\caption{A collection of random paths following the same model (top row: $T=T_{min}=18$; bottom row: $T =30$).}
\label{fig:random}
\end{figure}

3. The algorithm can be used in a generative mode to produce a collection of 
paths. In this case,  randomized actions could be taken following the same distributions available in the probability flow. For example, at time $t$, the next action can be sampled from the posterior  $p_{S_tA_t}(s_t,a_t) \propto f_{S_tA_t}(s_t,a_t) b_{S_tA_t}(s_t,a_t)$ as shown in  Figure \ref{fig:random}. When the time horizon is set to the minimum value (top row), the agent's different realizations are all best paths. Conversely, when $T$ exceeds the minimum time (bottom row),  the agent has  much more freedom to wander around and may take paths in different regions.  

\vspace{-0.9cm}
 \subsection{Steady state}

Forward and backward flows evolve according to the rules of probability propagation, explained in
Section \ref{sec:bayes} and shown in Figure \ref{fig:diff} for an example.  Even if they  are time-dependent, one may wonder whether 
the probability flow, more specifically the backward flow,  after a sufficient number of time steps, may reach a steady-state configuration. If such a convergence were verified, a unique configuration would allow simple  memorization of just one tensor and not the whole time-dependent flow. 
Unfortunately, the presence of obstacles and the reflection mechanism  causes the flows to oscillate in time and a steady-state configuration can be reached only in free areas. We have experimented using  the backward tensor (propagated back) at time $t=1$ throughout the whole time horizon. We have found that  the agent may fail to reach his goal, because it may undergo oscillatory behavior in the vicinity of obstacles.  

\section{Complexity and Memory}
\label{sec:complexity}

Tensor manipulation can become computationally heavy when we have to deal with large grids, for example, when the scene is modeled at high resolution. 
Complete memorization of the probability flow (forward, backward and posterior) for the overall time horizon $T$,  requires $\# real~value~memory~cells= 3NMn_AT$. 
The convolution-like computation of at every time step requires $\# operations=n_A^2 N M $
both for forward and backward flow. Posterior computation requires $n_A N M $ multiplications at every time step. Overall, the computational complexity can be estimated to be in the order of 
$O(n_A^2 N M T)$. To have an idea, a $100 \times 100$ grid with $n_a=9$ with a time horizon of $100$, requires approximately  27M locations and a total number of operations in the order of  81M (no graphics). To have a frame of reference, on a laptop with CPU i7-9750H for a grid 25x25 and $T= 64$,   our typical simulation runs in about 200 seconds, graphics included. 

\section{Relation to MDP and dynamic programming}
\label{sec:mdp}

There is clearly a connection between the Bayesian approach to path planning discussed here and dynamic programming that projects actions into the future to adopt control policies. 
In dynamic programming, there is a reward mechanism connected to states and actions at every time step. In the probabilistic approach presented in this paper, even if reward nodes  could be included in the graph (influence diagrams \cite{Influence2012}), they are left out for now and the only connection to the objectives are in the goal position and in the messages propagated back from the  final time $T$.  By definition of Bayesian networks, maximization of posterior probabilities corresponds to the maximization of the total likelihood
\begin{equation}
\begin{array}{l}
L=\pi_{S_1A_1}(s_1,a_1) \sum_{t=2}^{T-1}
p_{S_tA_t|S_{t-1}A_{t-1}}(s_t,a_t|s_{t-1},a_{t-1}) \\
p_{S_T|S_{T-1}A_{T-1}}(s_T|s_{T-1},a_{T-1}).
\end{array}
\end{equation}
A comparison with finite time MDP  with infinite time horizon, has been established by Toussaint, \cite{Touissaint2006} that considers a mixture of models with a finite random $T$, distributed according to a discount-like prior $P(T)=\gamma^T (1-\gamma)$.  By using a fixed reward $r=1$ at every time step, he establishes the equivalence of the likelihood maximization to the expected total reward. Many are the variations available for implementing  value iteration in the MDPs. For example, in path planning, details on how to introduce obstacles may also vary. However,  Attias' probability propagation idea on fixed time \cite{Attias2003}, on which we have built most  of our experiments, seems to be much more appealing for visualizing the directional flow: once established, the minimum time $T$ necessary to reach the goal, as we have suggested above, with our renormalization model for obstacle avoidance, the agent is guaranteed to reach his objective in minimum time.  

\begin{figure}[h]
\centering
\includegraphics[width=8cm]{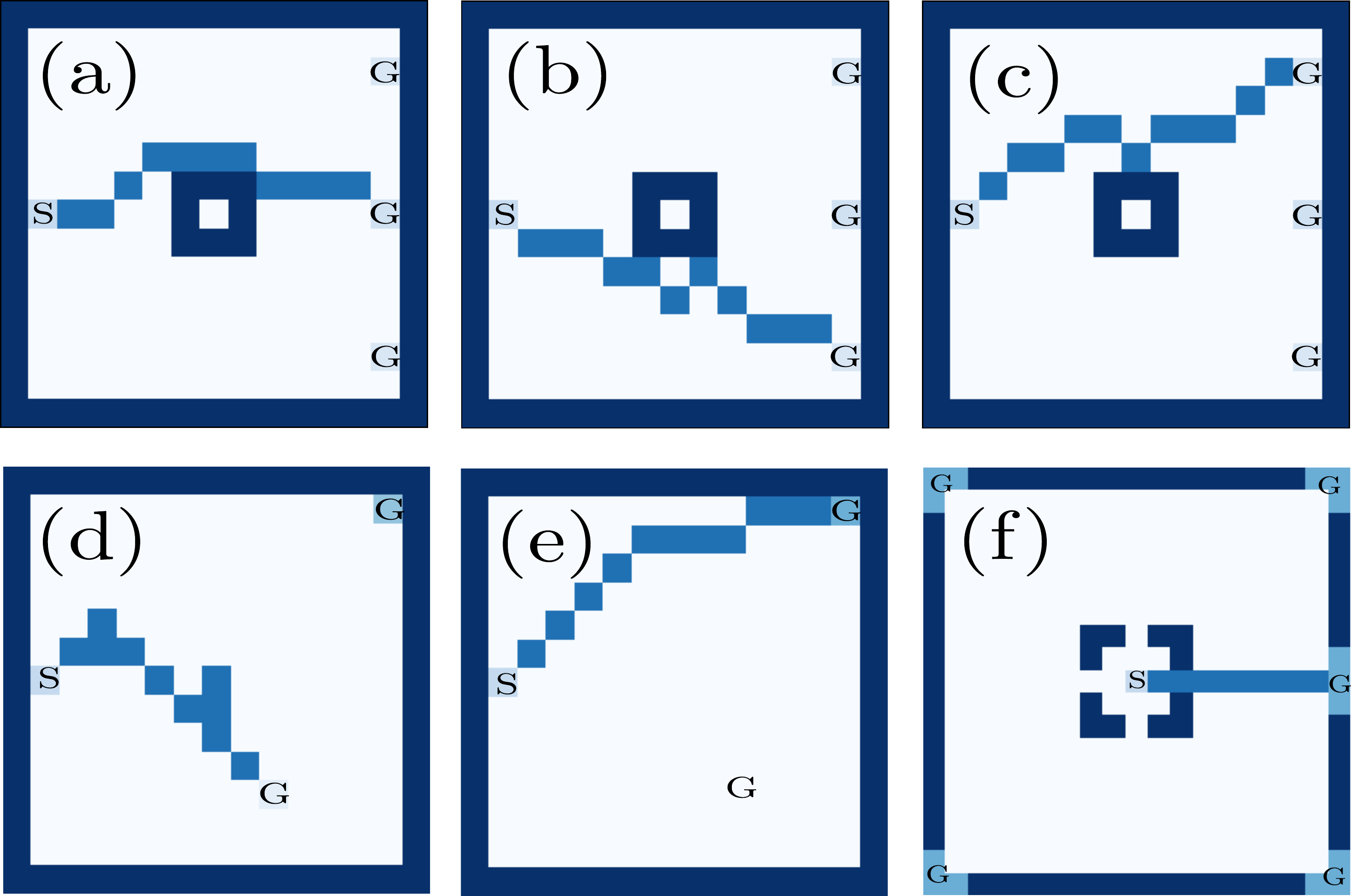} 
\caption{Paths obtained with multiple goals. In (a-c) there are three goals with the same backward probability and at the same distance form the start: the agents randomly chooses one of them. In (d) the two goals have backward probability 0.2 and 0.8, for the closer and the farthest respectively. The agents essentially ignores the farthest goal. In (e) we have the same situation of (d), but the backward probabilities are pushed to .0001 and .9999: here the agent does have enough ``attraction'' from the farthest goal and sometimes it goes there. Plot (f) is a grid with many doors as goals. Since here the backward probabilities are uniform, the agent goes toward the closest door at the closest point.}
\label{fig:multiple}
\end{figure}

\vspace{-.5cm}
\section{Multiple Goals}
\label{sec:multiple}

This probabilistic framework allows great flexibility in modeling the agent's motion. For example, there may be more than one target. Distributed goals can be introduced, by injecting at the end of the chain a backward message distributed over multiple locations. Figure \ref{fig:multiple} shows the results of simulations with an agent that has to reach more than one goal. Various scenarios can be considered: 

1. The time horizon $T$ is set to the minimum value for the farthest goal, and the backward distribution at $T$ is uniform on all the goals. In this case, the greedy algorithm G  always tends to reach the goal at  the minimum distance, because the likelihood is maximized. If the actions are randomized, all goals may be reached, but in the various realizations we have lower occurrence for the farthest ones. Figures \ref{fig:multiple}(a-c) show three occurrences for one agent and three goals in which all the goals are at the same distance and $T=T_{min }=13$. 

2. For an agent to reach the farthest goal more often, backward probabilities may be controlled. 
For example, in Figure \ref{fig:multiple}(d), we have set probabilities 0.2 and 0.8 on the two goals.  However, they are not sufficient to direct the agent toward the farthest goal often enough: the picture shows a path to the closest path, but in multiple trials we have never seen the farthest goal reached.  In Figure  \ref{fig:multiple}(e) the farthest goal has been assigned a probability of 0.9999  and  the closest 0.0001. The agent reaches the farthest goal more often and the picture shows such a captured realization. In the field theory analogy, the attraction force associated to the backward flow, for the farthest source, has to equalize, or overcome, the other. 

3. Multiple goals may be distributed on arbitrary locations in the grid to model the presence of doors or target areas. Figure \ref{fig:multiple}(f) shows a grid with many doors. In that case, the probabilities injected at the end are uniform and the agent tends to go towards the closest door. Unbalanced backward probabilities would produce other paths. 

Variation on this scenario, both on maps and goal distributions,  have been included in our experiments and are not reported here for brevity.   

\section{Multiple agents}
\label{sec:many}

\begin{figure*}[ht]
\centering
\includegraphics[width=16cm]{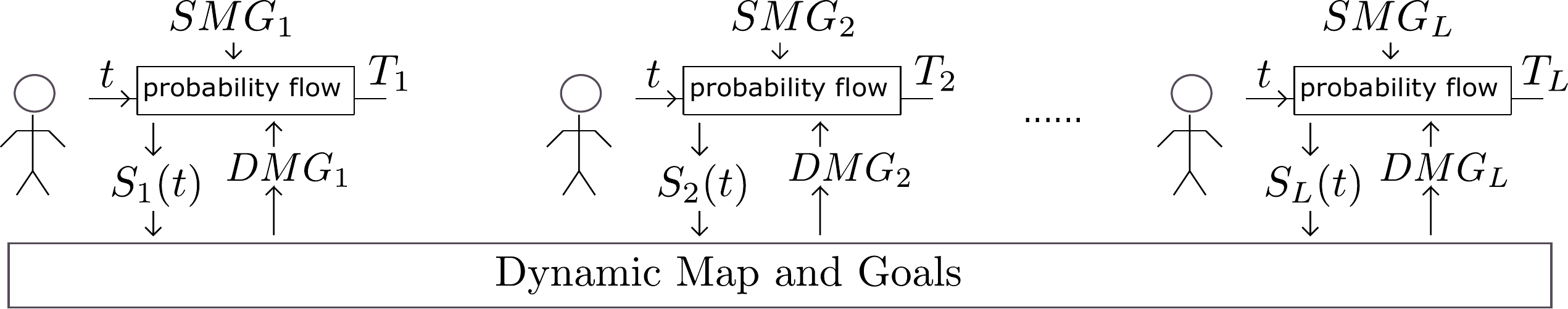} 
\caption{Interacting agents that keep their probability flow updated according to their Static Map and Goals (SMG) and their Dynamic Map and Goals (DMG). The agents, scheduled in sequence, have knowledge of the current state of the others,   that is translated in dynamically changing individual map and goals.}
\label{fig:Lagents}
\end{figure*}

We have started to use this probabilistic  framework to model much more complicated situations, such as those arising when there are multiple agents that have to reach different goals. Figure \ref{fig:Lagents} shows a diagram with $L$ agents that keep their probability flow updated according to both static and dynamically-changing map and goals. The agents are scheduled to act in a sequence (that can also be randomly permuted), and have their current state visible to the others.  This is our first cut to approach parallel behavior, by assuming that each agents acts independently from the others, but sees the others just as obstacles, and/or as goals. Clearly, the map and/or the goals for each agent change at every time step and the whole probability flow has to be recomputed dynamically. We have run many simulations with multiple agents and the emerging behaviors are amazingly close to what is observed on real scenarios! 

\begin{figure}[h]
\centering
\includegraphics[width=8cm]{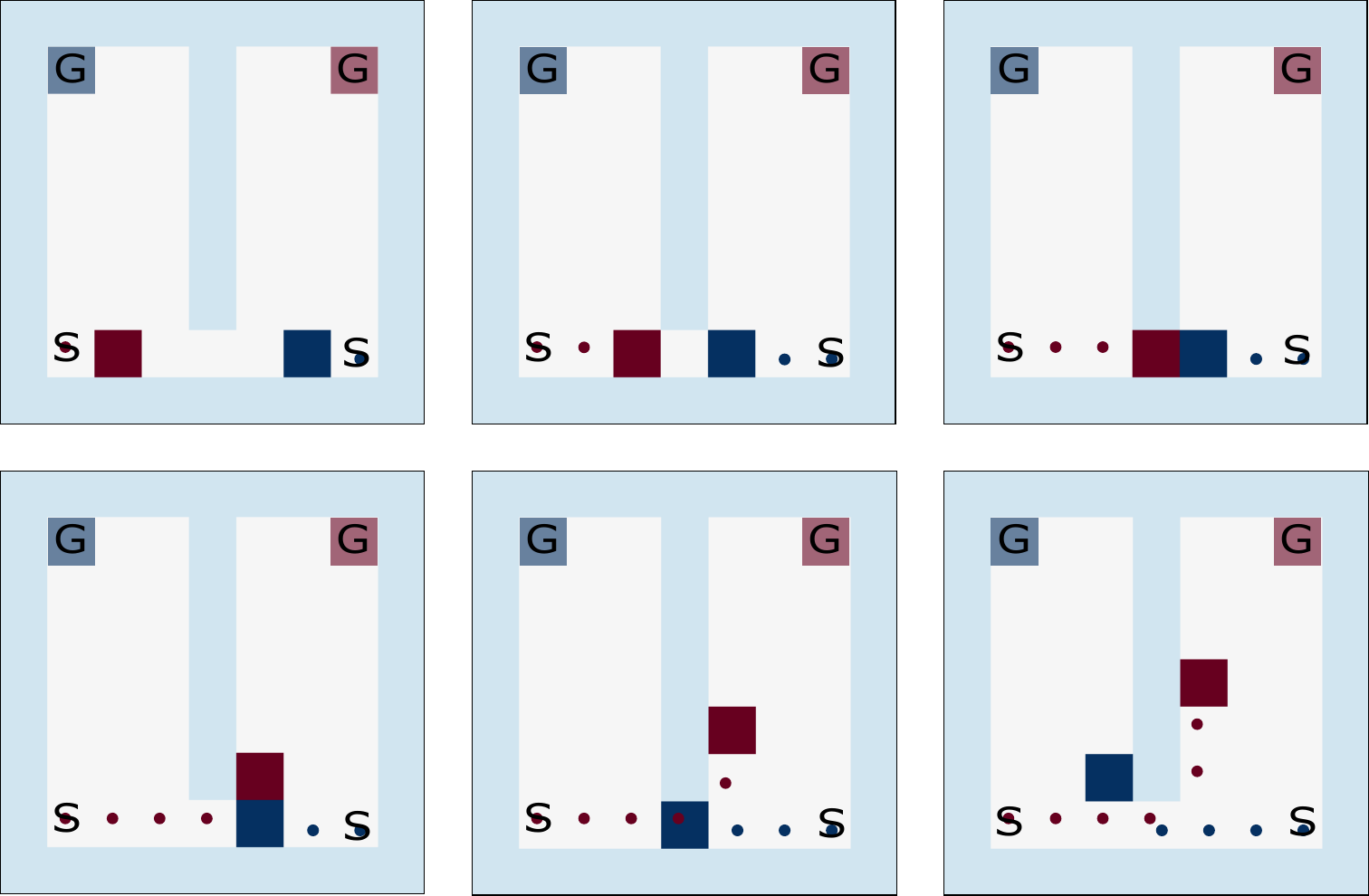} 
\caption{Two agents trying to reach their respective goal. A narrow passage does not allow simultaneous crossing. Therefore, since each agent sees the other as an obstacle, one of the agents waits for the passage to open, i.e. for a feasible solution to become available. }
\label{fig:2agents}
\end{figure}

Figure \ref{fig:2agents} shows a grid where there are two agents that are trying to reach their respective goals on a map with a narrow passage that does not allow simultaneous crossing. What happens running the two probability flows in parallel, is that when one of the two agents ``sees''  that the path becomes obstructed (no feasible solution exists at that time step to reach the goal), the agent  ``waits'' until a feasible solution becomes available. As pointed out in one of the above sections, when a solution is not available, the agent can adopt various strategies that go from staying still, or move randomly (in our case the agents wait). Since the grid is continuously changing and the agents' actions are scheduled in a sequence,  the probability flows will possibly change favorably and the agent  will be guided in the right direction.  

Agents could also be ``looking for  each other'' by simply setting each agent's position as the goal for one or more other agents. The results of these simulations will be reported elsewhere.    
 
\begin{figure*}[h]
\centering
\includegraphics[width=17cm]{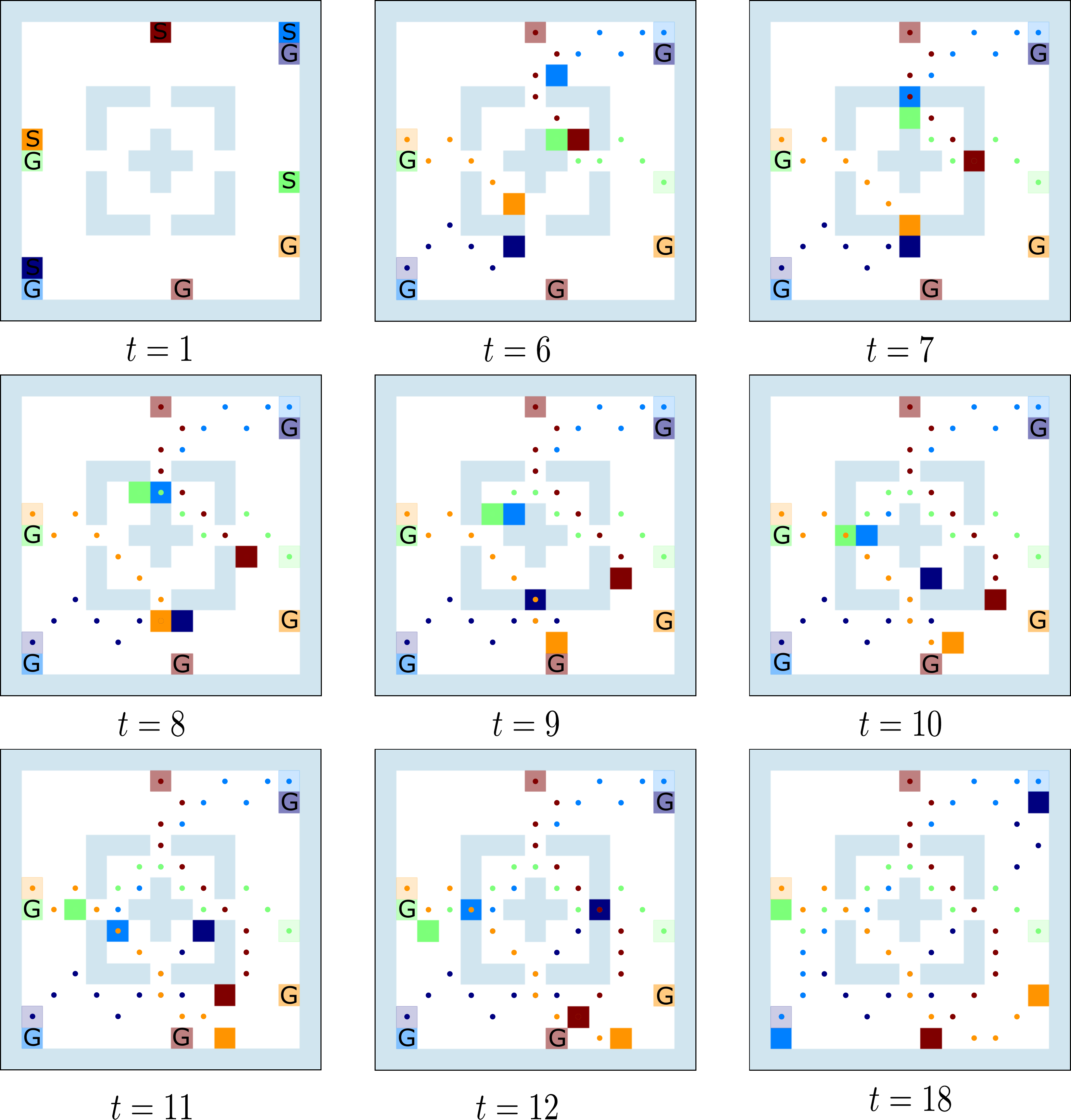} 
\caption{Five agents try to reach their respective goals (agents and goals are color coded: very light for the start and light for the goal). The picture shows 9 frames ($t=1,6,7,8,9,10,11,12,18$) of a rather complex behavior, where agents may have to cross narrow openings and avoid each other by waiting, changing paths, etc. Various behaviors are worth to note: (a) In frames $t=7, 8, 9$,  we can note a conflict between the dark-blue and the orange agents. The orange agent at $t=7$ is blocking the passage and the dark-blue agent chooses to go right at $t=8$ beginning a new feasible path. However, as soon as the passage opens, the dark-blue agent decides to go back at $t=9$, because that path is better than the one he had initiated. (b) In frame $t=7$, because of the blockage caused by the orange agent, the red agent decides to take an external path, rather than go trough the obstacles.  (c) In frame $t=10$, the green agent obstructs the passage for the blue agent that chooses first to go below at $t=11$. Then as soon as the passage opens, he goes back toward that passage at $t=12$, because it corresponds to a better path. 
}
\label{fig:nagents}
\end{figure*}

 In Figure \ref{fig:nagents}, there are five agents, each trying to reach its goal. Only nine frames are shown. The limitations of static images does not show the full potential of the model that causes the agents to avoid each other, to wait if necessary, to take different paths, etc. (see caption).
Animated plots will be made available on our website.  

\section{Conclusions and Future Work}
\label{sec:conc}

The probabilistic framework to model motion in complex scenarios is very promising because it accounts for  ``intelligent'' behaviors. We have shown how tensor messages  can be used to solve apparently very complex path planning problems. The forward and the backward flows, that are derived from the Bayesian formulation, are available to the agent  that has to undertake the proper actions to reach his final goal. It clearly demonstrates  the crucial role played by the backward flow, that being essentially the system running backward in time, provides guidance into the future. The backward flow resembles a vector field and the consequences of this findings are intriguing also in the context of free-energy models. 

 In this paper, we have first extended the single-agent framework to distributed goals, where, in realistic settings, the model can easily include doors, benches,  counters, food sources, etc.  Then, we have extended the model to account for multiple agents: the agents' Markov models  are run in parallel and, in our first approach to the problem,  each agent sees the others as moving obstacles (or goals). The emerging behaviors are very realistic and we believe that the consequences of  this powerful framework are yet to be explored. 
 
More work is in progress on the following: 1. Application of this framework to real scenes; 2. Inclusion of  learning; 3. Inclusion of dynamic rewards; 4. Partial knowledge of goals and obstacles; 5. Partially observable states.  



\bibliographystyle{IEEEtran}
\vskip 0.2in
\bibliography{ProbPropBib,ref}
\end{document}